\documentclass{article}
\usepackage{spconf,amsmath,graphicx,hyperref}
\usepackage{amssymb, amsthm}

\usepackage{multirow}
\usepackage{booktabs}
\usepackage{siunitx}
\usepackage{array} 
\usepackage{marvosym}

\title{MGKAN: Predicting Asymmetric Drug-Drug Interactions via a Multimodal Graph Kolmogorov-Arnold Network}
\name{Kunyi Fan$^{1,*}$ \qquad Mengjie Chen$^{1,2,*}$ \qquad Longlong Li$^{1}$ \qquad 
Cunquan Qu$^{2,\dagger}$\thanks{\footnotesize $^{*}$These authors contributed equally to this work. \protect\\ \hspace*{1.8em} $^{\dagger}$Corresponding author: Cunquan Qu (cqqu@sdu.edu.cn)
\protect\\ \hspace*{1.8em} \textbf{This paper has been accepted by ICASSP 2026.}
}}

\address{$^{1}$School of Mathematics, Shandong University, Jinan, China \\
         $^{2}$Data Science Institute, Shandong University, Jinan, China}
\begin{document}
%
\maketitle
\begin{abstract}
Predicting drug-drug interactions (DDIs) is essential for safe pharmacological treatments. Previous graph neural network (GNN) models leverage molecular structures and interaction networks but mostly rely on linear aggregation and symmetric assumptions, limiting their ability to capture nonlinear and heterogeneous patterns. We propose MGKAN, a Graph Kolmogorov-Arnold Network that introduces learnable basis functions into asymmetric DDI prediction. MGKAN replaces conventional MLP transformations with KAN-driven basis functions, enabling more expressive and nonlinear modeling of drug relationships. To capture pharmacological dependencies, MGKAN integrates three network views—an asymmetric DDI network, a co-interaction network, and a biochemical similarity network—with role-specific embeddings to preserve directional semantics. A fusion module combines linear attention and nonlinear transformation to enhance representational capacity. On two benchmark datasets, MGKAN outperforms seven state-of-the-art baselines. Ablation studies and case studies confirm its predictive accuracy and effectiveness in modeling directional drug effects. 
\end{abstract}
\begin{keywords}
Asymmetric drug-drug interactions, directed graph neural networks, Kolmogorov-Arnold networks
\end{keywords}
\section{Introduction}

\begin{figure}[h]
    \centering
    \includegraphics[width=0.975\columnwidth]{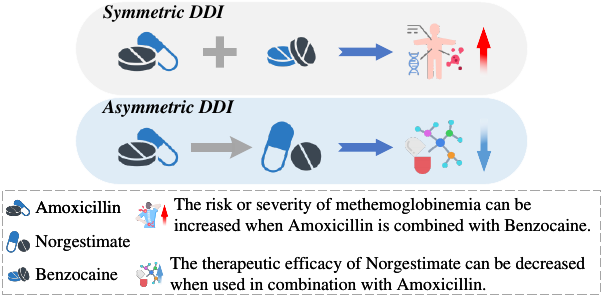}
    \caption{Examples of symmetric and asymmetric DDIs.}
    \vspace{-2mm}
    \label{fig:ADDI}
\end{figure}

Drug-drug interactions (DDIs) can compromise therapeutic efficacy or increase toxicity when drugs are administered together~\cite{zhao2024drug}. A critical but often underexplored property of DDIs is asymmetry—many interactions are directional, meaning the source drug significantly influences the effect or metabolism of the target drug, while the reverse interaction does not hold~\cite{zhang2023application}. As illustrated in Fig.~\ref{fig:ADDI}, the therapeutic efficacy of Norgestimate can be decreased when used in combination with Amoxicillin, whereas Norgestimate has no significant effect on the efficacy of Amoxicillin~\cite{simmons2018drug}. The directionality of DDIs plays a crucial role in polypharmacy, as it determines the optimal sequence of drug administration and supports safer clinical decision-making.

Despite its critical role in clinical decision-making, the directionality of DDIs is often overlooked by existing computational models, including traditional machine learning approaches~\cite{han2022review,wang2023comprehensive} and mainstream graph neural network approaches~\cite{li2024deep,zhao2024drug}, which typically treat interactions as undirected and fail to capture asymmetry. Recently, some studies have attempted to account for the asymmetry of DDIs~\cite{deng2024mavgae,zhang2024drgatan}. For instance, DGAT-DDI~\cite{feng2022directed} introduced a directed graph attention network to learn role-specific drug embeddings. However, most of these models aggregate information solely within the DDI network, overlooking high-order interaction patterns and failing to incorporate biochemical features from other modalities. These limitations prevent existing models from capturing the complex, role-specific mechanisms underlying asymmetric DDIs.

To address this issue, we propose MGKAN, a \textbf{M}ultimodal \textbf{G}raph \textbf{K}olmogorov-\textbf{A}rnold \textbf{N}etwork for asymmetric DDI prediction. MGKAN constructs three views—an asymmetric DDI network, a co-interaction network, and a biochemical similarity network—and applies direction-aware message passing with KAN to learn nonlinear, role-specific embeddings. A fusion module integrates multi-view information via linear attention and nonlinear transformation for final prediction. The key contributions of our research are as follows:

\begin{itemize}

    \item We construct three multi-view networks capturing direct interactions, topological relations, and biochemical features, which provide complementary information for drug embedding learning. 

    \item We introduce KAN into graph message passing to learn the source and target role-specific embeddings of drugs through learnable basis functions with direction-aware message passing to capture asymmetry more efficiently.

    \item We conduct a variety of experiments on different DDI datasets to demonstrate the validity and superiority of the proposed MGKAN model.

\end{itemize}

\section{Preliminaries}

\textbf{Asymmetric DDI Network.} We formally define the asymmetric DDI network as a directed graph $\mathcal{G}=(\mathcal{V},\mathcal{E})$, where $\mathcal{V}$ is the set of drugs and $\mathcal{E}\subseteq \mathcal{V}\times \mathcal{V}$ is the set of interactions. A directed edge $\langle u,v \rangle \in \mathcal{E}$ represents an interaction $u \rightarrow v$, where the source drug $u$ affects the target drug $v$.

\noindent\textbf{Asymmetric DDI Prediction.} Given $\mathcal{G}$ and drug features $X \in \mathbb{R}^{N \times d}$, we define two prediction tasks for a drug pair $(u,v)$: 
(1) \textbf{Specific-direction prediction}, which determines whether a directed interaction exists from drug $u$ to drug $v$; and 
(2) \textbf{Asymmetric-direction prediction}, which predicts the interaction into $u \to v$, $v \to u$, or no interaction. 

\noindent\textbf{Graph Kolmogorov-Arnold Network (GKAN).}
GKAN integrates KANs~\cite{liu2024kan} into GNNs by replacing MLPs with adaptive basis functions. For node $i$, at layer $l$, messages from neighbors $j\!\in\!\mathcal{N}(i)$ are first computed using learnable B-spline~\cite{de2024kolmogorov} basis functions $\phi(\cdot)$ and then aggregated with degree normalization:
\begin{equation}
    \phi(x_j^{(l)}) = \sum_{k} c_k B_k(x_j^{(l)}), \quad 
    \mathbf{m}_{j \to i}^{(l)} = \tfrac{\phi(x_j^{(l)})}{\sqrt{d_i d_j}},
\end{equation}
where $B_k$ are basis splines, $c_k$ are learnable coefficients and $d_i$, $d_j$ are the degrees of nodes $i$ and $j$, respectively. The aggregated message for node $i$ is $\mathbf{m}_i^{(l)}=\sum_{j\in\mathcal{N}(i)}\mathbf{m}^{(l)}_{j\to i}$, and the node embedding is updated by:
\begin{equation}
    x_i^{(l+1)}=\Phi(\mathbf{m}_i^{(l)}), \quad 
    \Phi(x)=\omega_b b(x)+\omega_s \phi(x),
\end{equation}
where $b(\cdot)$ is a standard activation (e.g., SiLU), and $\omega_b,\omega_s$ are learnable weights. In matrix form, it can be expressed as:
\begin{equation}
    X^{(l+1)} = \Phi ( \hat{A} \cdot \phi( X^{(l)} ) ),\quad
    \hat{A}=D^{-\frac{1}{2}} A D^{-\frac{1}{2}},
\end{equation}
where $A$ is the adjacency matrix and $D$ is the degree matrix.

\section{Methodology}
\begin{figure*}[ht]
    \centering
    \includegraphics[width=0.9\linewidth]{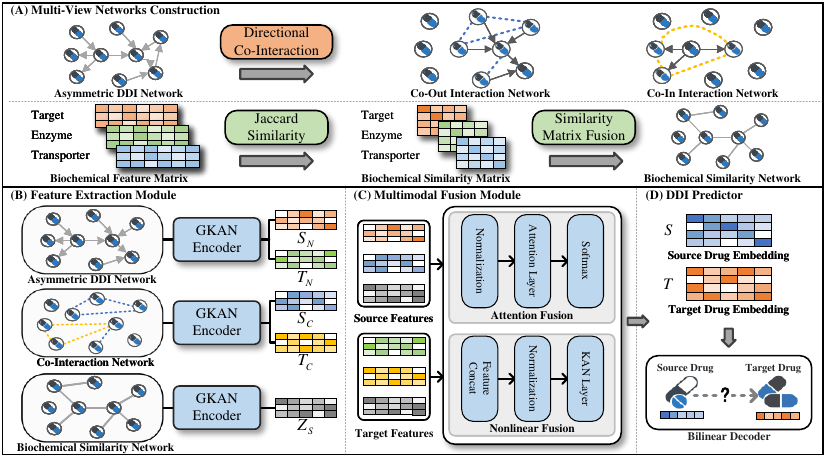}
    \vspace{-2mm}
    \caption{The overall framework of MGKAN.} 
    \vspace{-2mm}
    \label{fig:framework}
\end{figure*}

As shown in Fig.~\ref{fig:framework}, MGKAN first constructs three complementary views—an asymmetric DDI network, a co-interaction network, and a biochemical similarity network—to capture diverse structural and semantic information. The GKAN encoder then performs direction-aware message passing to learn source and target role embeddings. These embeddings are integrated by a fusion module that combines linear attention with nonlinear transformations. Finally, a bilinear decoder estimates the probability of asymmetric interactions.

\subsection{Feature Extraction Module}
\noindent\textbf{Feature extraction based on asymmetric DDI network.}
In asymmetric DDIs, drugs play distinct source and target roles depending on interaction direction. To distinguish this asymmetry, we separately aggregate different neighborhoods of drugs to obtain role-specific information. For a drug $u \in \mathcal{V}$, we define out-neighbors $\mathcal{N}_{out}$ and in-neighbors $\mathcal{N}_{in}$:
\[
\mathcal{N}_{out}(u)=\{v \mid \langle u,v\rangle \in \mathcal{E}\}, \quad 
\mathcal{N}_{in}(u)=\{v \mid \langle v,u\rangle \in \mathcal{E}\},
\]
with adjacency matrices $A_{out}$ and $A_{in}$, respectively. We independently aggregate information from $\mathcal{N}_{out}$ as the source embedding $S_N$ and from $\mathcal{N}_{in}$ as the target embedding $T_N$:
\begin{equation}
S_N=\Phi(\hat{A}_{out} \cdot \phi(X)), \quad 
T_N=\Phi(\hat{A}_{in} \cdot \phi(X)),
\end{equation}
where $\hat{A}_{out}$ and $\hat{A}_{in}$ are the degree-normalized adjacency matrices for out-neighbors and in-neighbors, respectively.

\noindent\textbf{Feature extraction based on co-interaction network.}
In asymmetric DDI networks, relying only on direct interactions limits topological information: network sparsity may leave some nodes without neighbors, and direct links cannot capture higher-order relations. To address this, we construct a co-interaction network to model higher-order associations.  Following~\cite{tong2020directed}, two drugs $u$ and $v$ are co-in neighbors if they are both interacted by the same drug $k$, and co-out neighbors if they both interact with the same drug $k$. The corresponding adjacency matrices are defined as follows:
\begin{equation}
\begin{aligned}
C_{in}(i,j)&=\sum_{k}\frac{A_{k,i}A_{k,j}}{\sum_{v}A_{k,v}}, \quad C_{out}(i,j)&=\sum_{k}\frac{A_{i,k}A_{j,k}}{\sum_{v}A_{v,k}},
\end{aligned}
\end{equation}
where $A$ is the adjacency matrix of $\mathcal{G}$. Then GKAN is applied to obtain source and target co-interaction embeddings:  
\begin{equation}
S_C=\Phi(\hat{C}_{out} \cdot \phi(X)), \quad 
T_C=\Phi(\hat{C}_{in} \cdot \phi(X)),
\end{equation}
where $\hat{C}_{out}$ , $\hat{C}_{in}$ are degree-normalized adjacency matrices of the co-interaction network. \textbf{Supplementary File} provides theoretical analysis of second-order co-interaction kernels.

\noindent\textbf{Feature extraction based on similarity network.}
To model biochemical influences on DDIs, we extract three biochemical features for each drug: targets, enzymes, and transporters, encoded as binary vectors. Pairwise similarities are computed using the Jaccard similarity:
\begin{equation}
Jaccard(d_u,d_v)=\frac{|d_u \cap d_v|}{|d_u \cup d_v|},
\end{equation}
where $d_u$ and $d_v$ are feature vectors of drugs $u$ and $v$.  
This produces three symmetric similarity matrices: target ($B_t$), enzyme ($B_e$), and transporter ($B_p$). We obtain the fused similarity matrix by element-wise addition $B = B_t \oplus B_e \oplus B_p$ , which serves as the adjacency matrix of the biochemical similarity network. Finally, GKAN aggregates information on this network to obtain similarity embeddings:
\begin{equation}
Z_S=\Phi(\hat{B} \cdot \phi(X)),
\end{equation}
where $\hat{B}$ is the degree-normalized similarity matrix.

\subsection{Multimodal Fusion Module}
After extracting embeddings from multi-view networks, we design a complementary fusion module combining attention-based linear fusion and KAN-based nonlinear fusion to enhance representation learning.

\noindent\textbf{Linear fusion via attention.}
Let $F_i$ denote the embedding of the $i$-th view, and compute the attention coefficient $\alpha_i$ as:
\begin{equation}
w_i=q^T \tanh(WF_i+b), \quad 
\alpha_i=\frac{\exp(w_i)}{\sum_k \exp(w_k)},
\end{equation}
where $q$, $W$, and $b$ are learnable parameters. Distinct coefficients $\alpha_*$ and $\beta_*$ are assigned to the source and target embeddings, respectively. The fused embeddings are:
\begin{equation}
\begin{aligned}
S^{\text{Attn}} &= \alpha_N S_N+\alpha_C S_C+\alpha_S Z_S, \\
T^{\text{Attn}} &= \beta_N T_N+\beta_C T_C+\beta_S Z_S.
\end{aligned}
\end{equation}

\noindent\textbf{Nonlinear fusion via KAN.}
While the attention module assigns weights to different embeddings, it essentially performs a linear combination. To capture nonlinear interactions, the concatenated embeddings are transformed by KAN:
\begin{equation}
S^{\text{NL}}=\Phi_{out}(S_N \| S_C \| Z_S), \quad 
T^{\text{NL}}=\Phi_{in}(T_N \| T_C \| Z_S).
\end{equation}

Finally, the two embeddings are concatenated to form the final source embedding $S$ and target embedding $T$:
\begin{equation}
S=S^{\text{Attn}} \| S^{\text{NL}}, \quad 
T=T^{\text{Attn}} \| T^{\text{NL}}.
\end{equation}

\subsection{Asymmetric DDI Prediction and Loss Function}
A bilinear decoder estimates asymmetric DDI probabilities:
\begin{equation}
\hat{y}_{u \to v}=\sigma(S_u^T M T_v),
\end{equation}
where $\sigma$ is the sigmoid function and $M$ is a learnable weight matrix.  
The loss function $\mathcal{L}$ is then defined as:
\begin{equation}
\mathcal{L}=-\frac{1}{\lvert \mathcal{D} \rvert}\sum_{\langle u,v\rangle\in \mathcal{D}}
\,
[ \small y_{\small u \to v}\log\hat{y}_{\small u \to v}+(1-y_{\small u \to v})\log(1-\hat{y}_{\small u \to v}) ],
\end{equation}
where $\mathcal{D}$ consists of both positive and sampled negative pairs from the training set.

\section{Experiments}
We evaluate MGKAN on two benchmark datasets against several state-of-the-art baselines. The experimental setup and additional details are provided in the \textbf{Supplementary File} at \url{https://github.com/KYFAN-02/MGKAN}.

\begin{table*}[ht]
\centering
\small
\renewcommand{\arraystretch}{0.8}
\caption{Performance comparison of MGKAN in \textbf{Task 1}, the highest value in each column is highlighted in bold.}
\setlength{\tabcolsep}{1.45mm}
\begin{tabular}{l*{8}{c}}
\toprule
\multirow{2}{*}{\textbf{Method}} & \multicolumn{4}{c}{\textbf{DS 1}} & \multicolumn{4}{c}{\textbf{DS 2}} \\
\cmidrule(lr){2-5} \cmidrule(l){6-9}
 & \textbf{AUROC} & \textbf{AUPRC} & \textbf{ACC} & \textbf{F1} & \textbf{AUROC} & \textbf{AUPRC} & \textbf{ACC} & \textbf{F1} \\
\midrule
Gravity VGAE~\cite{salha2019gravity} & $88.54_{\pm0.01}$ & $85.26_{\pm0.03}$ & $81.66_{\pm0.03}$ & $82.20_{\pm0.03}$ & $87.90_{\pm0.01}$ & $85.75_{\pm0.01}$ & $79.93_{\pm0.02}$ & $80.57_{\pm0.03}$ \\
S/T VGAE~\cite{salha2019gravity} & $92.14_{\pm0.08}$ & $91.51_{\pm0.10}$ & $84.13_{\pm0.13}$ & $83.72_{\pm0.22}$ & $91.26_{\pm0.04}$ & $90.75_{\pm0.07}$ & $83.23_{\pm0.12}$ & $82.94_{\pm0.26}$ \\
DiGAE~\cite{kollias2022directed} & $94.84_{\pm0.03}$ & $93.73_{\pm0.06}$ & $91.15_{\pm0.04}$ & $91.29_{\pm0.03}$ & $90.57_{\pm0.02}$ & $88.44_{\pm0.05}$ & $85.91_{\pm0.03}$ & $86.14_{\pm0.02}$ \\
DirGNN~\cite{rossi2024edge} & $97.82_{\pm0.02}$ & $97.38_{\pm0.04}$ & $93.24_{\pm0.03}$ & $93.22_{\pm0.03}$ & $95.55_{\pm0.02}$ & $94.99_{\pm0.03}$ & $89.18_{\pm0.03}$ & $89.13_{\pm0.04}$ \\
MAVGAE~\cite{deng2024mavgae} & $92.53_{\pm0.06}$ & $91.78_{\pm0.06}$ & $84.39_{\pm0.07}$ & $84.21_{\pm0.06}$ & $91.57_{\pm0.08}$ & $91.57_{\pm0.08}$ & $83.29_{\pm0.44}$ & $82.81_{\pm0.77}$ \\
DRGATAN~\cite{zhang2024drgatan} & $87.30_{\pm0.05}$ & $86.72_{\pm0.05}$ & $79.82_{\pm0.08}$ & $80.71_{\pm0.08}$ & $83.51_{\pm0.05}$ & $83.88_{\pm0.05}$ & $74.48_{\pm0.06}$ & $76.85_{\pm0.05}$ \\
DGAT-DDI~\cite{feng2022directed} & $98.55_{\pm0.03}$ & $98.20_{\pm0.04}$ & $94.69_{\pm0.12}$ & $94.67_{\pm0.14}$ & $96.65_{\pm0.02}$ & $95.98_{\pm0.03}$ & $90.91_{\pm0.04}$ & $91.02_{\pm0.06}$ \\
MGKAN & $\mathbf{99.08}_{\pm\mathbf{0.01}}$ & $\mathbf{98.94}_{\pm\mathbf{0.02}}$ & $\mathbf{95.05}_{\pm\mathbf{0.14}}$ & $\mathbf{94.90}_{\pm\mathbf{0.16}}$ & $\mathbf{98.09}_{\pm\mathbf{0.02}}$ & $\mathbf{97.62}_{\pm\mathbf{0.03}}$ & $\mathbf{93.04}_{\pm\mathbf{0.09}}$ & $\mathbf{92.93}_{\pm\mathbf{0.11}}$ \\
\bottomrule
\end{tabular}
\label{tab:Task_1}
\end{table*}

\subsection{Performance Comparison}
We evaluate MGKAN on specific-direction (Task 1) and asymmetric-direction (Task 2) prediction using the DS 1 and DS 2 datasets from DrugBank~\cite{knox2024drugbank} under a transductive 8:1:1 random link split using five-fold cross-validation with 1:1 negative sampling from unknown pairs. Full Task 2 results and a runtime analysis are provided in the \textbf{Supplementary File}, showing that MGKAN consistently outperforms baselines on both tasks while maintaining competitive efficiency.

As shown in Table~\ref{tab:Task_1}, MGKAN consistently outperforms all baselines on Task 1. While directed graph learning methods exhibit strong structural modeling capabilities, they achieve lower AUROC and F1, indicating that structural information alone is insufficient for accurate asymmetric DDI prediction. On the other hand, existing asymmetric DDI baselines primarily rely on the DDI network, limiting their ability to capture complex interaction patterns. MGKAN addresses these limitations by integrating multiple views—including co-interaction and similarity networks—which provide complementary topological and semantic information. For instance, drugs without direct interactions can still be indirectly connected through co-interaction networks, while the similarity view introduces auxiliary biochemical features. This multi-view architecture enables MGKAN to generate more expressive and discriminative embeddings than models restricted to a single view.

\subsection{Ablation Study}

\begin{figure}[htbp]
    \centering
    \includegraphics[width=0.975\columnwidth]{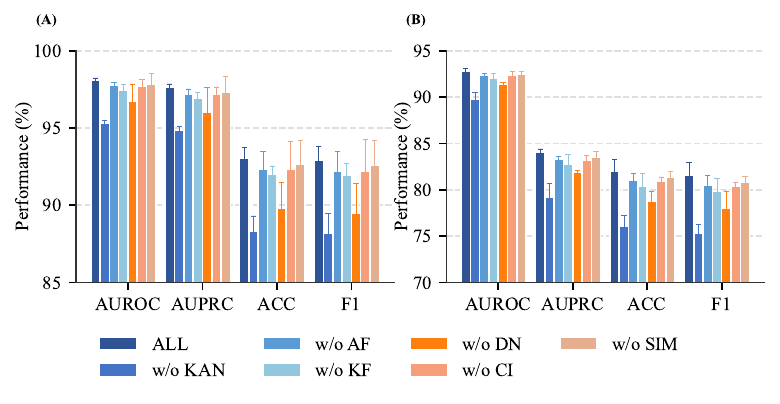}
    \vspace{-2mm}
    \caption{Ablation study results of MGKAN and its variants. 
    \textbf{(A)} Performance on \textbf{Task 1}. \textbf{(B)} Performance on \textbf{Task 2}.} 
    \vspace{-2mm}
    \label{fig:ablation_study}
\end{figure}

We conduct an ablation study to assess the contribution of each module in MGKAN. As shown in Fig.~\ref{fig:ablation_study}, we evaluate six variants: \textbf{w/o KAN} (replace KAN layers with MLPs), \textbf{w/o AF} (remove attention fusion), \textbf{w/o KF} (remove KAN-based fusion), \textbf{w/o DN} (no asymmetric DDI view), \textbf{w/o CI} (no co-interaction view), and \textbf{w/o SIM} (no similarity view).

The ablation study yields three key insights consistent with the MGKAN design. First, removing KAN causes the largest performance drop, confirming its central role in modeling nonlinear and directional patterns. Second, both fusion modules are important; notably, eliminating KAN-based fusion results in greater degradation, highlighting the value of nonlinear fusion and the complementarity of the two mechanisms. Third, all three views contribute: the asymmetric DDI network captures direct interactions, while co-interaction and similarity views provide additional topological and biochemical information. These results validate our design, and embedding visualizations in the \textbf{Supplementary File} further illustrate MGKAN’s expressiveness.

\subsection{Case Study}
To evaluate MGKAN’s ability to identify unobserved interactions, we perform a transductive prediction on DS 1 and validate the top 10 predicted pairs using the DrugBank interaction checker. As shown in Table~\ref{tab:case_study}, seven were asymmetric interactions, two symmetric interactions, and one unrecorded, demonstrating MGKAN’s potential for novel DDI discovery.

\begin{table}[htbp]
\renewcommand{\arraystretch}{0.8}
\small
\centering
\caption{Top 10 asymmetric DDIs predicted by MGKAN.}
\begin{tabular}{llll}
\toprule
Rank & Source drug & Target drug & Direction \\
\midrule
1 & Calcium carbonate & Dolutegravir & \checkmark \\
2 & Fluvoxamine & Mirtazapine & \checkmark \\
3 & Promethazine & Metyrosine & $\times$ \\
4 & Vemurafenib & Miglitol & \checkmark \\
5 & Lidocaine & Arformoterol & \checkmark \\
6 & Labetalol & Clenbuterol & \checkmark \\
7 & Leflunomide & Methotrexate & $\times$ \\
8 & Canagliflozin & Pioglitazone & \checkmark \\
9 & Argatroban & Glyburide & N/A \\
10 & Digoxin & Deslanoside & \checkmark \\
\bottomrule
\end{tabular}
\label{tab:case_study}
\end{table}

\vspace{-2mm}

\section{Conclusion}
This paper proposed MGKAN, a multi-view framework for asymmetric DDI prediction that integrates direction-aware message passing and nonlinear fusion via KAN.  Experiments on multiple datasets show that MGKAN effectively captures interaction directionality and outperforms existing methods. Future work will extend MGKAN to multi-relation DDI prediction and incorporate richer drug modalities.

\section*{Acknowledgment} This work was supported by the National Natural Science Foundation of China under Grant 12471488 and Grant 12231018.

\bibliographystyle{IEEEbib}
\bibliography{strings,new_refs}

\end{document}